\documentclass[10pt,twocolumn,letterpaper]{article}

\usepackage{iccv}              % To produce the CAMERA-READY version
\usepackage{xcolor}
\usepackage{algorithm}
\usepackage{algorithmic}
\usepackage{amsmath}
\usepackage{amssymb}
\usepackage{multirow}
\usepackage{xcolor}
\usepackage{pifont}
\usepackage{booktabs}
\usepackage{graphicx}
\definecolor{iccvblue}{rgb}{0.21,0.49,0.74}
\usepackage[pagebackref,breaklinks,colorlinks,allcolors=iccvblue]{hyperref}

\def\paperID{0000} % *** Enter the Paper ID here
\def\confName{ICCV}
\def\confYear{2025}

\title{\raisebox{-0.5\baselineskip}{\includegraphics[width=0.08\textwidth]{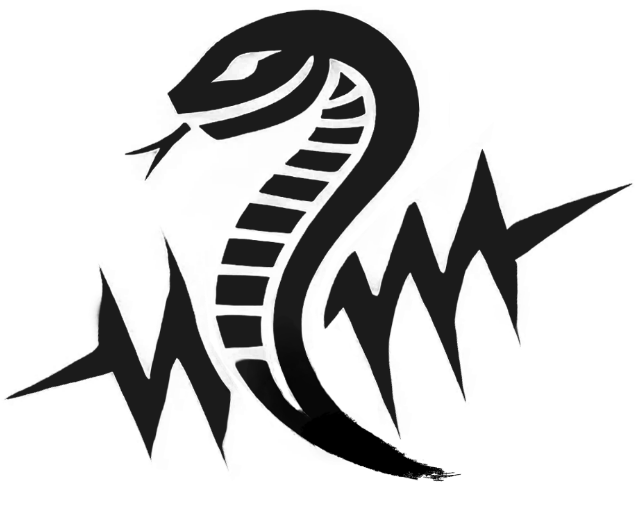}} Efficient Spiking Point Mamba for Point Cloud Analysis}

\author{
Peixi Wu\textsuperscript{1}, 
Bosong Chai\textsuperscript{2},
Menghua Zheng\textsuperscript{3},
Wei Li\textsuperscript{1}, 
Zhangchi Hu\textsuperscript{1}, \\
Jie Chen\textsuperscript{1},
Zheyu Zhang\textsuperscript{1},
Hebei Li\textsuperscript{1}\thanks{Corresponding authors.}, 
Xiaoyan Sun\textsuperscript{1}\footnotemark[1] \\
\\
\textsuperscript{1}University of Science and Technology of China \\ 
\textsuperscript{2}Zhejiang University  \quad
\textsuperscript{3}Tsingmao Intelligence \\  
\tt\small \{wupeixi, lihebei, sunxiaoyan\}@ustc.edu.cn 
}

\begin{document}
\maketitle
\begin{abstract}

Bio-inspired Spiking Neural Networks (SNNs) provide an energy-efficient way to extract 3D spatio-temporal features.
However, existing 3D SNNs have struggled with long-range dependencies until the recent emergence of Mamba, which offers superior computational efficiency and sequence modeling capability. 
% However,  the straightforward adoption of Mamba does not achieve satisfactory performance for point cloud analysis.
% This research aims to bridge the performance gap between ANNs and SNNs in point cloud analysis. 
In this work, we propose Spiking Point Mamba (SPM), the first Mamba-based SNN in the 3D domain.
Due to the poor performance of simply transferring Mamba to 3D SNNs, SPM is designed to utilize both the sequence modeling capabilities of Mamba and the temporal feature extraction of SNNs.
Specifically, we first introduce Hierarchical Dynamic Encoding (HDE), an improved direct encoding method that effectively introduces dynamic temporal mechanism, thereby facilitating temporal interactions.
Then, we propose a Spiking Mamba Block (SMB), 
which builds upon Mamba while learning inter-time-step features and minimizing information loss caused by spikes. 
Finally, to further enhance model performance, we adopt an asymmetric SNN-ANN architecture for spike-based pre-training and finetune.
Compared with the previous state-of-the-art SNN models, SPM improves OA by +\textbf{6.2}\%, +\textbf{6.1}\%, and +\textbf{7.4}\% on three variants of ScanObjectNN, and boosts instance mIOU by +\textbf{1.9}\% on ShapeNetPart. 
Meanwhile, its energy consumption is at least \textbf{3.5}$\times$ lower than that of its ANN counterpart. 
% The code will be made publicly available.

% at least 6.4x less than its ANN counterpart.

% which extends Mamba by learning inter-time-step features and minimizing information loss caused by spikes. 

% with a significant improvement of +\textbf{6.2}\%, +\textbf{6.1}\%, +\textbf{7.4}\% OA on three variants of ScanObjectNN and +\textbf{1.9}\% Ins. mIOU on ShapeNetPart. Meanwhile, its energy consumption is at least \textbf{3.5}$\times$ lower than that of its ANN counterpart, PointMamba.
% ------------Bosong-------------

% Recent research has explored the application of SNNs in 3D point cloud analysis, aiming to provide low-power solutions for various scenarios. This paper introduces Spiking Point Mamba (SPM), the first Mamba-based SNN framework designed specifically for 3D point cloud analysis.

% To enhance sequence modeling capabilities, we propose Hierarchical Dynamic Encoding (HDE), an improved direct encoding method that efficiently introduces dynamic temporal mechanisms into SPM. Additionally, we introduce the Spiking Mamba Block (SMB), which builds upon the powerful modeling capabilities of Mamba while learning features across time steps and minimizing information loss caused by spikes.

% Furthermore, we pioneer the introduction of pre-training in the 3D SNN domain, leveraging a heterogeneous SNN-ANN architecture based on Mamba to boost the performance of SPM across various datasets and tasks. Experimental results demonstrate significant performance improvements in downstream tasks, validating the effectiveness of combining Mamba with 3D SNNs.
\end{abstract}    
\section{Introduction}

% bosong ------------------------
% Bio-inspired Spiking Neural Networks (SNNs) are considered as the third generation of neural networks~\cite{maass1997networks}. The sparse spike-driven mechanism in SNNs means that neurons do not perform computations at every time step, significantly reducing the consumption of computational resources. 

\begin{figure}
    \centering
    \includegraphics[width=1.0\linewidth]{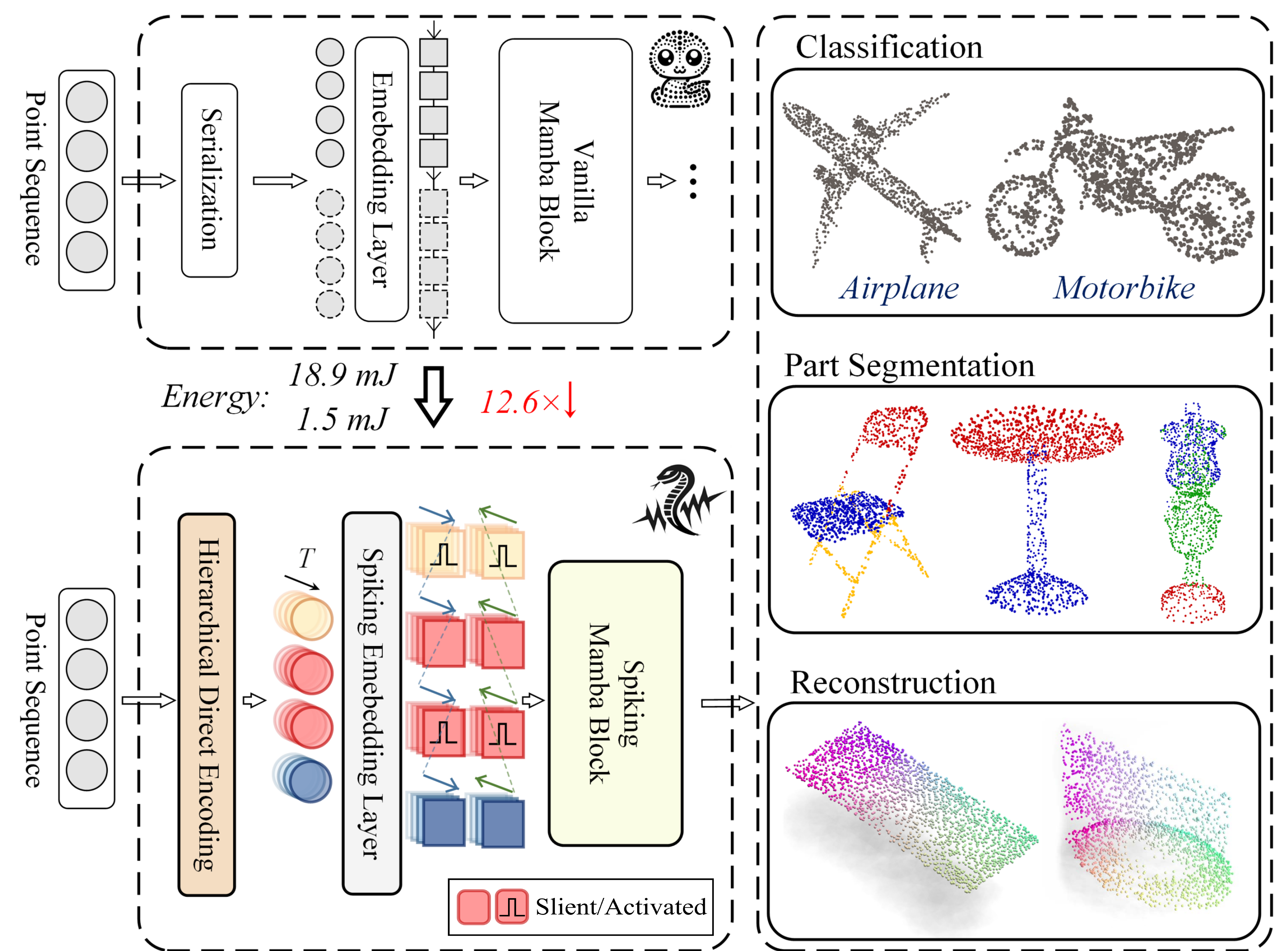}
    \caption{
    The SNN-adapted design of Spiking Point Mamba, compared to PointMamba~\cite{liang2024pointmamba}, serves as an efficient backbone for various 3D point cloud analysis tasks, including object classification, part segmentation, and reconstruction.}
    % SNN-adapted design of Spiking Point Mamba compared with PointMamba and its visualizations of  downstream tasks.}
    % Spiking Point Mamba适应SNN的架构设计及其与PointMamba对比。以及所做的三个下游点云分析任务。}
    \label{fig:Head}
\end{figure}

Bio-inspired Spiking Neural Networks (SNNs) offer a promising energy-efficient paradigm for 3D point cloud analysis, leveraging event-driven computation and intrinsic spatio-temporal dynamics~\cite{schuman2022opportunities,roy2019towards}. Despite recent advances in MLP-based and Transformer-based SNN architectures~\cite{ren2024spiking,wu2024point,qiu2024efficient}, critical limitations persist: (1) Existing models struggle to capture long-range dependencies (LRDs) in irregular point sequences; (2) Static temporal encoding methods~\cite{zhou2024spikformer,ouyang4706194spiking} fail to exploit the dynamic nature of spike-driven temporal features; (3) Information degradation during spike-driven computation remains unaddressed~\cite{zhou2024qkformer}, particularly when integrating modern sequence modeling techniques.

The recent success of Mamba architectures in 3D vision~\cite{liang2024pointmamba,han2024mamba3d} provides new opportunities. Their selective state space mechanisms enable more effective LRD modeling~\cite{gu2023mamba}. However, naively adapting Mamba to SNNs faces two challenges: (1) Temporal complexity mismatch between continuous state transitions and discrete spike events, and (2) Information density disparity (30\% fewer activations per timestep compared to ANNs~\cite{zhong2024spike,zhou2024qkformer}).

To bridge this gap, we propose Spiking Point Mamba (SPM), the first Mamba-based SNN framework that synergizes energy-efficient spike computation with powerful sequence modeling. First, we introduce an improved direct encoding called Hierarchical Dynamic Encoding (HDE) to SPM. Most of the high-performance SNN studies~\cite{zhou2024spikformer,ouyang4706194spiking,qiu2024efficient} are based on direct encoding, repeating the input along the time dimension to generate a static encoding sequence. However, it fails to fully exploit the potential of SNNs in extracting the temporal feature effectively~\cite{zhang2024tc,yin2024dynamic}.
To this end, HDE introduces dynamics in the early unstable and late redundant stages of farthest point sampling (FPS) while preserving the key information at each time step in the middle stage. 
HDE transforms the static point cloud into a hierarchical, dynamic event-based representation, facilitating better feature interaction in SPM.

Then, we propose Spiking Mamba Block (SMB) as the core component of SPM to leverage the capacity of both Mamba and SNNs.
Although Mamba demonstrates strong performance in sequence modeling~\cite{gu2023mamba,dao2024transformers}, its original architecture is not fully compatible with the spike-based input~\cite{zhong2024spike,shen2024spikingssms}.
To this end, we customize the Gate branch and the SSM feature extraction branch of SMB. 
The design not only preserves the modeling capacity of Mamba, but also enhances the dynamic interactions across time steps while minimizing the information loss caused by spike. 
To boost the performance of SPM across various datasets and tasks, we pretrain SPM through a heterogeneous SNN-ANN architecture, effectively leveraging the modeling strengths of 3D ANNs to enhance the generalization of SPM. 

% We introduce the Spiking Mamba Block (SMB) as the core of SPM, integrating the strengths of Mamba and SNNs. While Mamba excels in sequence modeling~\cite{gu2023mamba,dao2024transformers}, its original design is incompatible with spike-based inputs~\cite{zhong2024spike,shen2024spikingssms}. To address this, we modify the Gate branch and SSM feature extraction branch in SMB, preserving Mamba's ability to model internal dependencies while enhancing temporal dynamics and reducing spike-induced information loss. Additionally, we pretrain SPM using a heterogeneous SNN-ANN architecture, leveraging the representational power of 3D ANNs to improve generalization across datasets and tasks.

In summary, our main contributions are as follows:

\begin{itemize}
    \item  We build Spiking Point Mamba (SPM), the first energy-efficient Mamba-based 3D SNN framework, and introduce spike-based pre-training to the 3D SNN domain.
    % \item  We design Hierarchical Dynamic Encoding (HDE), an improved direct encoding method that efficiently introduces dynamic temporal mechanism, enhancing temporal interactions in SNNs.
    \item We propose Hierarchical Dynamic Encoding (HDE), an improved direct encoding method that enhances temporal interactions via dynamic temporal mechanism.
    \item We introduce Spiking Mamba Block (SMB), which builds upon the modeling capabilities of Mamba while learning 
    features across time steps effectively. 
    \item Extensive experimental results demonstrate that SPM achieves significant performance improvements in various downstream task.
\end{itemize}

\section{Related Works}

\noindent \textbf{State Space Models.}
Inspired by continuous state space models in control systems, recent research has shown that SSM~\cite{fuhungry,mehtalong}, as a promising alternative to sequence models such as RNNs and Transformers, can effectively model LRDs. To maintain performance while reducing computational costs, HTTYH~\cite{gutrain}, DSS~\cite{gupta2022diagonal}, and S4D~\cite{gu2022parameterization} propose using diagonal matrices within S4~\cite{gu2021efficiently}. S5~\cite{smithsimplified} introduce parallel scanning and MIMO SSM. Recently, Mamba~\cite{gumamba} introduces the selective SSM mechanism, achieving linear-time inference and efficient training. 
Meanwhile, various variants of SSMs~\cite{orvieto2023resurrecting} have been successfully applied to many domains involving  audio~\cite{goel2022s,erol2024audio}, video~\cite{li2024videomamba,yang2024vivim} and vision~\cite{nguyen2022s4nd,liu2025vmamba,vim}. 

\noindent \textbf{Deep Learning for Point Cloud.}
Point cloud analysis follows two main approaches: projecting raw point clouds onto voxel grids or images~\cite{maturana2015voxnet,shi2020pv,li2020end} , and directly processing raw point cloud. The success of transformers in deep learning has driven a shift from MLP-based~\cite{ma2022rethinking,qi2017pointnet,qi2017pointnet++} to transformer-based methods~\cite{wu2022point,park2022fast,wu2024point}, but their quadratic complexity increases computational costs, posing challenges for long-sequence point clouds.
Recently, Mamba has emerged as a more efficient alternative with linear complexity and strong sequence modeling~\cite{li2024videomamba,vim}. By effectively capturing long-range dependencies (LRDs), Mamba-based methods~\cite{liu2024mamba4d,zhang2024point,zhang2024voxel,liu2024point,liang2024pointmamba} provide a more efficient and  scalable solutions for point cloud analysis.

\noindent \textbf{SNN Training and Architecture Design.}
The primary challenge in training SNNs is the non-differentiable spike function~\cite{qiu2024efficient}. To address this, recent research focuses on improving training strategies and architectures. Some studies~\cite{hao2023reducing,bu2023optimal} concentrate on ANN-to-SNN conversion~\cite{hu2023fast,wang2023masked}, which transforms trained ANNs into equivalent SNNs using neuron equivalence. However, this approach requires long time steps and increases energy consumption. Alternatively, other works directly train SNNs using surrogate gradients~\cite{guo2023rmp}, striving to make computations spike-driven as much as possible~\cite{yao2024spike}, thereby enabling low energy use and high performance with shorter time steps. Our architecture follows this direct training paradigm.

% The primary challenge in training SNNs is the non-differentiable nature of the spike function~\cite{qiu2024efficient}. To address this, current research focuses on improving training strategies and architectures. 
% Some studies~\cite{hao2023reducing,bu2023optimal} concentrate on ANN-to-SNN conversion~\cite{hu2023fast,wang2023masked}, which transforms trained ANNs into equivalent SNNs using neuron equivalence. However, this method requires long simulation time steps and increases energy consumption. 
% As a result, another methods focus on direct training using surrogate gradients~\cite{guo2023rmp}, striving to make computations spike-driven as much as possible~\cite{yao2024spike}, 
% thereby enabling low energy consumption and high performance with shorter time steps. We design our novel SNN architecture based on the latter approach.
% --------------------------------------------------------
\section{Preliminaries}
\subsection{LIF Neuron}
% The Leaky Integrate-and-Fire (LIF) spiking neuron is the most popular neuron to balance bio-plausibility and computing complexity.

\begin{figure*}
    \centering
    \includegraphics[width=1.0\linewidth]{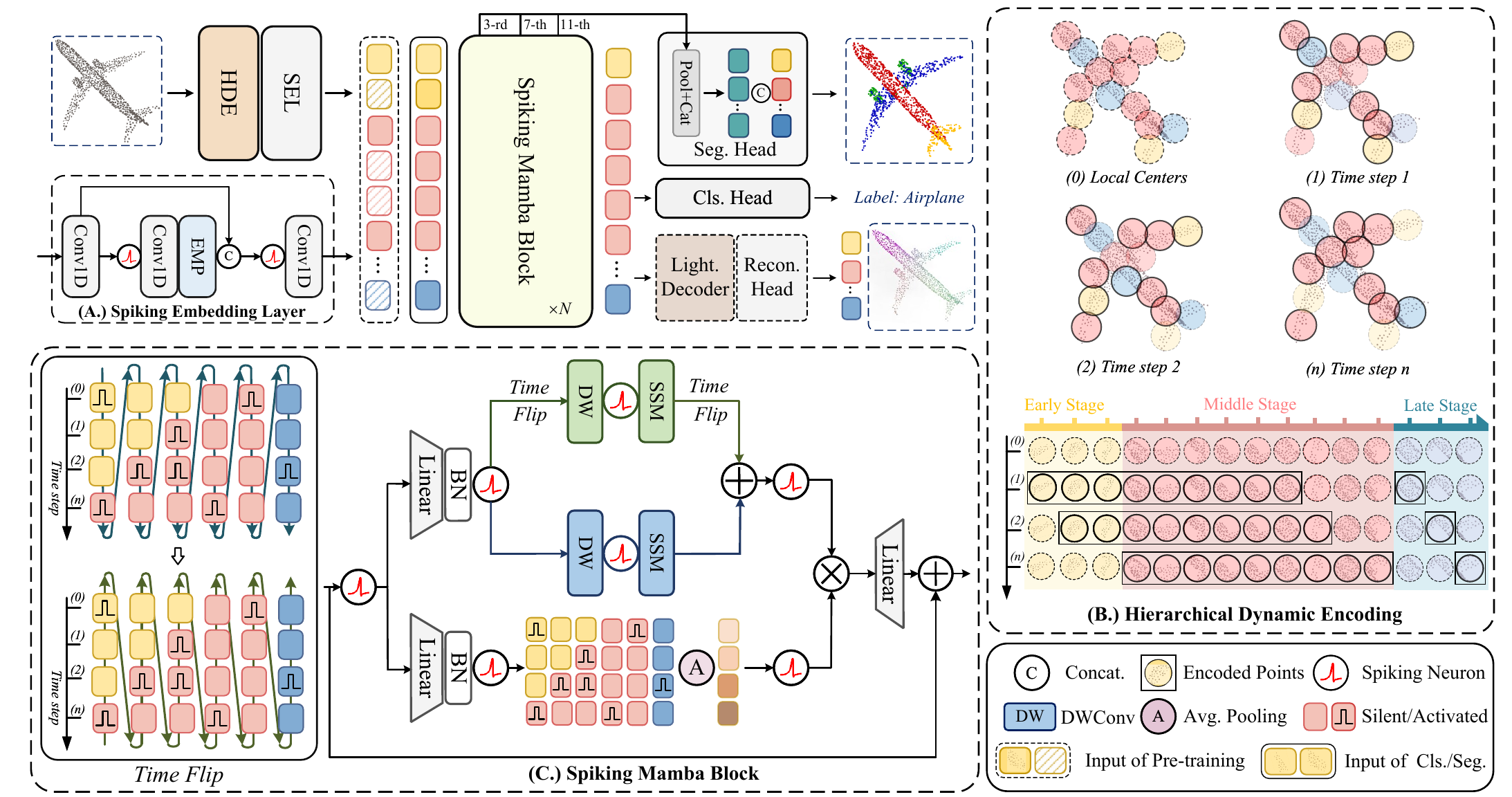}
    \caption{The overview of Spiking Point Mamba (SPM), which consists of Hierarchical Dynamic Encoding (HDE), Spiking Embedding Layer (SEL) for adaptive learning, Spiking Mamba Block (SMB) for feature interaction. For downstream tasks, we design Classification and Segmantation Head, while for pre-training, we design  Reconstruction Head with a lightweight decoder.}
    \label{fig:Main}
\end{figure*}

The Leaky Integrate-and-Fire (LIF) neuron, a simplified biological neuron model, captures the "leaky-integrate-fire-reset" process~\cite{sedighi2024visual}. It is the most popular neuron due to its balance of bio-plausibility and computing complexity. Given time step 
$t$, the LIF neuron is formulated by following equations:
\begin{align}
\label{eq:1}
&\mathbf{H}_{t} = f(\mathbf{V}_{t-1}, \mathbf{X}_{t}) \\ \label{eq:2}
&\mathbf{S}_{t} = \Theta(\mathbf{H}_{t} - V_{th}) \\ \label{eq:3}
&\mathbf{V}_{t} = \mathbf{H}_{t} \cdot (1 - \mathbf{S}_{t}) + V_{r} \cdot \mathbf{S}_{t}
\end{align}

\noindent where \( \mathbf{H}_t \) and \( \mathbf{V}_t \) are defined as the membrane potential after neuronal dynamics and after spike triggering, with \( \mathbf{X}_t \) as the external input and \( V_{th} \) as the firing threshold. The output spike \( \mathbf{S}_t \) follows ~\cref{eq:2}, where $\Theta(\cdot)$ denotes the Heaviside function. 
After firing, \( \mathbf{V}_t \) resets to \( V_{r} \) as shown in ~\cref{eq:3},
while ~\cref{eq:1} models the leaky-integrate process, where \( f(\cdot) \) governs decay and input accumulation.
We uniformly adopt the LIF neuron for $\mathcal{SN}$ in the following sections.
% \begin{equation}
%     H_{t}=V_{t-1}+\frac{1}{\tau}(-(V_{t-1}-V_{reset})+X_{t}).
% \end{equation}

% where $\tau$ denotes the time constant and $V_{reset}$ is the reset  membrane potential.

\subsection{State Space Model}

Drawing inspiration from control theory, the State Space Model (SSM) represents a continuous system that maps an input state \( \mathbf{x} \) to an output \( \mathbf{y} \) through an implicit latent state \( \mathbf{h} \). For a discrete input sequence
 $\mathbf{x}_{1:L}$, through certain discretization rule the SSMs can be defined by:
 \begin{align}
&\mathbf{h}_{k}=\bar{\mathbf{A}}\mathbf{h}_{k-1}+\bar{\mathbf{B}}\mathbf{x}_{k} \\
&\mathbf{y}_{k}=\mathbf{C}\mathbf{h}_k + \mathbf{D}\mathbf{x}_k
\end{align}

\noindent where \( \mathbf{C} \) is a projection parameter, and \( \mathbf{D} \) is a residual connection. The parameters \( \bar{\mathbf{A}} \) and \( \bar{\mathbf{B}} \) are discretized using the zero-order hold (ZOH) rule, introducing the timescale parameter \( \mathbf{\Delta} \). The SSMs are applied separately to each channel with data-independent parameters, limiting flexibility in sequence modeling. Recently, Mamba introduces the selective SSM, where \( \mathbf{B} \), \( \mathbf{C} \), and \( \mathbf{\Delta} \) depend on the input, improving the sequence modeling. In the following sections, both SSM and Mamba refer to the S6 model.
% \clearpage

% \section{Method}
% \label{sec:Method}

% --------------------------------------------------------
\section{Method}
\subsection{Main Architecture}
We propose Spiking Point Mamba (SPM), the first Spiking Neuron Network framework based on Mamba architecture, as shown in \cref{fig:Main}. First, Hierarchical Dynamic Encoding (HDE) is applied to the point cloud for time-dynamic encoding, generating tokenized representations. These tokens are then mapped to semantic features via Spiking Embedding Layer (SEL). Spiking Mamba Block (SMB) further  models both inter-token and inter-time-step interactions. Finally, the output is passed to different heads for various downstream tasks.

% \vspace{3cm}
% --------------------------------------------------------
\subsubsection{Hierarchical Dynamic Encoding}

Most SNN studies rely on direct encoding by repeating the input along the time dimension, limiting the temporal feature extraction due to its static nature. We introduce Hierarchical Dynamic Encoding (HDE), which treats each time step as richer input, introducing temporal dynamics. 
However, given the temporal correlation of neurons, it is essential to preserve key information at each time step while introducing dynamic changes to avoid distortion. 

Meanwhile,
considering the characteristics of farthest point sampling (FPS), it can be divided into three stages: early, middle, and late. The early stage is unstable primarily due to the random initial selection, and the middle stage stabilizes effectively capturing the skeletal structure of the point cloud, while the late stage may introduce redundancy or noise because of the sufficient sampling, as further validated in \cref{sec:behavior}. Hence, we process these three stages hierarchically to introduce dynamic changes.

As shown in \cref{algo:1}, HDE first downsamples the original point set $\mathbf{P}$ to obtain the sampled set $\mathbf{S}$, which is then divided into three stages: early, middle, and late, with $L$, $M$, and $R$ points respectively.
In the early and middle stages, we dynamically select $F$ points from $\mathbf{S}$ using a shifting sampling strategy with a step size of $l$. This dynamic mechanism ensures diversity in the initial sampling distribution while retaining key structural information. Due to the instability of randomly selected initial points within a limited range, $L$ remains constant, while $l$ decreases with time, referred to as \textit{Finite Forward Sliding}.
In the late stage, $r$ points are dynamically selected from the remaining $R$ points as a memory pool for each time step, referred to as \textit{Infinite Backward Extension}, where $R$ increases with time, while $r$ remains constant.
Finally, the point sets from \textit{Finite Forward Sliding} and \textit{Infinite Backward Extension} are merged to generate the encoded point matrix $\mathbf{E}$.
Together, the entire HDE process is  obversed in \cref{fig:Main}.

% --------------------------------------------------------
\subsubsection{Spiking Embedding Layer}
After Hierarchical Dynamic Encoding (HDE), the K-Nearest Neighbors (KNN) algorithm is applied to each encoded center point to find its $K$ nearest neighbors, treated as tokens $\mathbf{T}$. These tokens are then mapped by Spiking Embedding Layer (SEL) to obtain higher-dimensional semantic features, which are further processed  by SMB.
% , which are further processed by SMB to extract deeper inter-token and inter-timestep relationships.

Specifically, SEL projects the tokens to derive the membrane potential \( \mathbf{U} \) with preliminary learned deeper semantic information. 
Subsequently, an element-wise MaxPooling (EMP) is applied across all tokens to capture the local context of each token, which is then concatenated with $\mathbf{U}$ as an additional feature. 
This combined representation is further processed through a series of fully connected layers and spiking neurons to obtain the final embedded feature membrane potential \( \mathbf{U}' \). Finally, the entire representation is obtained by adding the potential with its positional embedding. Together, SEL can be formulated as:  
\begin{align}
&\mathbf{U} \hspace{1pt} = \hspace{0.8pt}{\rm{MLP}}(\mathcal{SN}({\rm{MLP}}(\mathbf{T}))) \\
&\mathbf{G} \hspace{0.5pt} = \hspace{0.5pt} {\rm{EMP}}(\mathbf{U}), \hspace{1pt} {\rm{RPE}}={\rm{MLP}}(\mathbf{E})  \\
&\mathbf{U}' = {\rm{MLP}}(\mathcal{SN}(Concat(\mathbf{U},\mathbf{G}))) \\
&\mathbf{U}_0 = \mathbf{U}'+{\rm{RPE}} 
\end{align}

\noindent where $\rm{MLP}$ is composed of a conv layer and a batch normalization 
 layer. $\rm{RPE}$ means relative position embedding. 
 $\mathbf{T}$ and $\mathbf{U}_0$ are features in ${\mathbb{R}}^{T \times E \times K\times C}$ and ${\mathbb{R}}^{T \times E \times C'}$, representing the tokens indexing based on KNN from $\mathbf{E}$ and the final output of SEL  respectively.

% --------------------------------------------------------
\subsubsection{Spiking Mamba Block}

Spiking Mamba Block (SMB) is the core component of the Spiking Point Mamba (SPM), integrating the strengths of both Mamba and SNNs. 
SMB processes both the temporal dynamics introduced by HDE and the interactions between membrane potential tokens. Similar to previous research, Spiking Point Mamba iteratively applies the SMB \( N \) times, refining the input membrane potential \( \mathbf{U}_0 \) at each block. The output of each SMB is defined as \( \mathbf{U}_n \).

% It not only retains the long-range token dependency handling capability of Mamba but also ensures the time feature extraction ability and low energy consumption characteristic of SNNs. 
% In the spirit of \textit{Occam's Razor}, SMB is designed without complex or elaborate plugins, thus facilitating deployment of SPM on neuromorphic hardware.

\begin{algorithm}[t]
\caption{Hierarchical Dynamic Encoding}
\label{algo:1}
\begin{algorithmic}[1]
\STATE \textbf{Result:} Encoded point matrix $\mathbf{E} \in {\mathbb{R}}^{T \times E \times C}$
\STATE \textbf{Input:} Point set $\mathbf{P}$, Sampled Points  $S$, Time Step  $T$
\STATE     \hspace{28pt}       Early Stage $L$, Mid Stage $M$, Late Stage $R$

\STATE \textcolor{gray}{\# \textit{Sampling}}
\STATE \(  \mathbf{S} \gets \text{FPS}(\mathbf{P}, S) \) 
\STATE \textcolor{gray}{\# \textit{Encoding}}
\STATE \(  l = \lfloor L/T \rfloor \), \(  r = \lfloor R/T \rfloor \)
\STATE \( M \gets M + L \), \( F \gets E - r \)

\FOR{$i = 1, 2, 3, \ldots, T$} 
    \STATE \textcolor{gray}{\# \textit{Finite Forward Sliding}}
    \STATE \( \mathbf{F}_i \gets \{ s_{j} \mid i \cdot l \leq j < F + i \cdot l \} \)
    
    \STATE \textcolor{gray}{\# \textit{Infinite Backward Extension}}
    \STATE \( \mathbf{B}_i \gets \{ s_{j} \mid M + i \cdot r \leq j < M + (i+1) \cdot r \} \)
    
    \STATE \( \mathbf{E}_i \gets \mathbf{F}_i \cup \mathbf{B}_i \)
\ENDFOR

\end{algorithmic}
\label{alg_1}
\end{algorithm}
Specifically, \( \mathbf{U}_n \) is first processed by a spiking neuron to generate spike \( \mathbf{S}_n \), enabling sparse synaptic accumulation in subsequent operations. SMB consists of two branches: SSM branch and Gate branch.
For SSM branch, it first encodes \( \mathbf{S}_n \) with fully connected layers and neurons to extract the intermediate feature spike \( \mathbf{S}_n' \).
 Unlike previous works, reversing the token dimension in \( \mathbf{S}_n' \) is not meaningful, as it is a sparse spike matrix with less information. Instead, we reverse only the time dimension to help SSM branch learn dynamic relationships between time steps. After timestep-wise interactions, \( \mathbf{S}_n'' \) is obtained through a spiking neuron. 
 For Gate branch, it first maps \( \mathbf{S}_n \) to a higher dimension to generate a gating matrix \( \mathbf{Z}_n \). However, element-wise multiplication between \( \mathbf{Z}_n \) and \( \mathbf{S}_n'' \) leads to significant information loss and disrupts token relationships. To address this, we apply element-wise average pooling (EAP) on \( \mathbf{Z}_n \) along the token dimension, preserving important feature dimensions while maintaining token relationships. Finally, \( \mathbf{U}_{n+1} \) is obtained through a linear layer and residual connection. Together, SMB can be formulated as:
\begin{align}
&\mathbf{S}_n = \mathcal{SN}(\mathbf{U}_n) \\
&{\mathbf{S}_n'}=\mathcal{SN}({\rm{MLP}}(\mathbf{S}_n)) {\mathbf{Z}_n}=\mathcal{SN}({\rm{MLP}}(\mathbf{S}_n))  \\
&\mathbf{U}_n', \mathbf{U}_{t}' = {\rm{SSM}}(\mathcal{SN}({\rm{DWConv}}(\mathbf{S}_n', \mathbf{S}_{t}'))) \\
&\mathbf{S}_n'' = \mathcal{SN}(\mathbf{U}_n'+\mathbf{U}_{t}') \circ \mathcal{SN}({\rm{EAP}}(\mathbf{Z}_n)) \\
&\mathbf{U}_{n+1} = {\rm{MLP}}(\mathbf{S}_n'') + \mathbf{U}_n
\end{align}

\noindent where \( \mathbf{S}_{t}' \) and \( \mathbf{U}_{t}' \) represent the time dimension reversal of \( \mathbf{S}_{n}' \) and its output through the SSM module, while \( \circ \) denotes the Hadamard product between the spike tensors. $n=1,2,...,N$ represents the layer number.

% --------------------------------------------------------
\subsection{Spike-based Pre-training Modeling}
Inspired by the success of pre-training in 2D SNNs, we introduce a masked modeling pre-training framework into 3D SNNs for robust point cloud feature representations, as shown in \cref{fig:Main}. A key challenge is to maximize the effectiveness of spike-based pre-training. To this end, we utilize an asymmetric SNN-ANN heterogeneous encoder-decoder architecture based on SMB and Mamba Block.

% Specifically, the encoder stacks SMB \(N\) times, while the decoder uses unidirectional SSM stacked \(N_d\) times for lightweight ANN decoding. $N_d$ is always less than $N$. Chamfer Distance is used as the reconstruction loss to recover the original point cloud. This design preserves the asymmetry between the encoder and decoder, speeding up pre-training. It also couples SNN and ANN, leveraging the ANN decoder’s modeling capabilities to enhance the SNN encoder's representation capacity while maintaining low energy consumption during inference. The spike-based pre-training paradigm enables SPM to achieve superior performance across datasets and tasks. Together, the spike-based pre-training modeling can be expressed as:
% \begin{align}
% & \mathbf{T}_{v}' = \mathcal{E}_{\rm{SMB}}(\mathbf{T}_{v}) \\
% & \mathbf{H}_{v}, \mathbf{H}_{m} = \mathcal{D}_{\rm{SSM}}(\mathbf{T}_{v}', \mathbf{T}_{m}) \\
% & \mathbf{P}_{m} = \mathcal{H}(\mathbf{H}_{m})
% \end{align}
% where $\mathbf{T}_{v}$, $\mathbf{T}_{v}'$ and $\mathbf{T}_{m}$表示掩码后可见的Tokens及其经过Encoder后，掩码后不可见的Tokens.
% $\mathbf{H}_{v}$ and $\mathbf{H}_{m}$ 表示解码器解码得到的.
% $\mathcal{E}_{\rm{SMB}}$ and $\mathcal{D}_{\rm{SSM}}$.
% $\mathcal{H}$.
% \begin{figure}
%     \centering
%     \includegraphics[width=1.0\linewidth]{fig/Main_v2.pdf}
%     \caption{The overview of Spiking Point Transformer (SPT)}
%     \label{fig:Pre}
% \end{figure}

Specifically, the encoder consists of \(N\) stacked SMBs, while the decoder utilizes a unidirectional SSM that is stacked \(N_d\) times for lightweight ANN decoding, where \(N_d < N\). This design preserves the asymmetry between the encoder and decoder, accelerating the pre-training. It also decouples SNN and ANN, leveraging the ANN decoder’s modeling capability to enhance the SNN encoder’s representation capacity while ensuring low energy consumption during inference.  Chamfer Distance is used as the reconstruction loss to recover the original point cloud.
Together, the spike-based pre-training modeling can be formulated as follows:
\begin{align}
& \mathbf{T}_{v}' = \mathcal{E}_{\rm{SMB}}(\mathbf{T}_{v}) \\
& \mathbf{H}_{v}, \mathbf{H}_{m} = \mathcal{D}_{\rm{SSM}}(\mathbf{T}_{v}', \mathbf{T}_{m}) \\
& \mathbf{P}_{m} = \mathcal{H}(\mathbf{H}_{m})
\end{align}

\noindent where \(\mathbf{T}_{v}\) and \(\mathbf{T}_{m}\) denote the visible and masked tokens after token masking, respectively, while \(\mathbf{T}_{v}'\) represents the encoded features of the visible tokens. \(\mathbf{H}_{v}\) and \(\mathbf{H}_{m}\) are the decoded features. \(\mathbf{P}_{m}\) denotes the reconstructed point cloud. \(\mathcal{E}_{\rm{SMB}}\) and \(\mathcal{D}_{\rm{SSM}}\) represent the encoder stacked with SMB and decoder stacked with SSM, respectively, and \(\mathcal{H}\) is the reconstruction head. 
% \(\mathbf{P}_{m}\) denotes the reconstructed point cloud, which is optimized using Chamfer Distance to ensure faithful recovery of the original structure.

% --------------------------------------------------------
\subsection{Theoretical Energy Consumption}
In this section, we investigate energy efficiency of our SPM architecture. SPM converts matrix multiplication into sparse addition.  Specifically, unlike ANNs where almost all FLOPs are multiply-accumulate (MAC) operations, SNNs use sparse spikes with only a subset of neurons activated for sparse accumulation (AC). According to the research~\cite{brette2005adaptive}, a 32-bit floating-point consumes 4.6$pJ$ for a
 MAC operation and 0.9$pJ$ for an AC operation, namely $E_{{\rm{MAC}}}=4.6pJ$ and $E_{{\rm{AC}}}=0.9pJ$.
% Therefore, SNNs are more energy-efficient than ANNs during inference.

The energy consumption of SPM primarily involves SEL and SMB. The SEL module consists of the initial floating-point MAC operations and the subsequent spike-based synaptic accumulation operations. The energy consumption can be expressed as:
\begin{align}
\hspace{-6.5pt} E_{{\rm{SEL}}} = E_{{\rm{MAC}}}\cdot FL_{conv}^{\mathbf{T},\mathbf{E}}+E_{{\rm{AC}}}\cdot FL_{conv}^{\mathbf{U},\mathbf{G}}\cdot T \cdot f_{conv}^{\mathbf{U},\mathbf{G}}
\end{align}
\noindent where \( f_{conv}^{\mathbf{U},\mathbf{G}} \) represent the firing rates of $\mathbf{U}$,$\mathbf{G}$ after $\mathcal{SN}$, while \( FL_{conv}^{\mathbf{T}, \mathbf{E}} \), \( FL_{conv}^{\mathbf{U}, \mathbf{G}} \) denote the FLOPs of corresponding conv layers of $\mathbf{T}$, $\mathbf{E}$, $\mathbf{U}$ and $\mathbf{G}$.

In the SMB module, all conv layers transmit spikes and performs AC operations, ensuring efficient and event-driven processing. SSM involves most AC operations with a small number of MAC operations introduced due to the floating-point multiplications between \( \Bar{\mathbf{A}} \), \( \mathbf{C} \), and hidden state \( \mathbf{h} \). The energy consumption can be given by:
\begin{align}
&\hspace{10pt} E_{\rm{SSM}} = E_{{\rm{MAC}}}\cdot FL_{1}^{n}+E_{{\rm{AC}}}\cdot FL_{2}^{n} \cdot T \cdot f_{2}^n \\
&\hspace{10pt} E_{{\rm{SMB}}} = E_{{\rm{SSM}}} + E_{{\rm{AC}}}\cdot FL_{conv}^{n} \cdot T \cdot f_{conv}^n
\end{align}

\noindent where \( f_{conv}^n \) and \( f_2^n \) denote the firing rates of the input spikes to different conv layers and SSM in $n$-th SMB respectively.
\( FL_{conv}^n \), \( FL_1^n \) and \( FL_2^n \) represent the FLOPs of different conv layers, floating-point multiplication, and spike-based multiplication in $n$-th SSM respectively. $T$ represents the time step.

% \newpage

\section{Experiments}
\subsection{Experimental Settings}

\begin{table*}[t]
  \small
  \centering
  \tabcolsep=0.17cm
  % \resizebox{\linewidth}{!}{
    \begin{tabular}{ccccccccc}
    \toprule
    \multicolumn{1}{c}{\multirow{2}[2]{*}{Architecture}} & 
    \multicolumn{1}{c}{\multirow{2}[2]{*}{Methods}} & 
    \multicolumn{1}{c}{\multirow{2}[2]{*}{Params}} & 
    \multicolumn{1}{c}{\multirow{2}[2]{*}{$T$}} & 
    \multicolumn{3}{c}{ScanObjectNN} & 
    \multicolumn{2}{c}{ModelNet40} \\ 
    \cmidrule(lr){5-7}\cmidrule(lr){8-9}        
    &      &  &  & OBJ-BG & OBJ-ONLY & PB-T50-RS & w/o Vote & w/ Vote \\  
    \midrule
    \multirow{4}[0]{*}{ANN} & PointNet \cite{qi2017pointnet} & 3.5  & -  & 73.3 & 79.2  & 68.0  & 89.2  & - \\ 
     & PointNet++ \cite{qi2017pointnet++} & 1.5  & - & 82.3  & 84.3  & 77.9  & 90.7  & - \\ 
     % & PoinCNN \cite{li2018pointcnn} & 0.6  & - & 86.1 & 85.5 & 78.5  & 92.2 & - \\ 
     & PointMLP \cite{ma2022rethinking} & 13.2  & - & -  & - & 85.4  & 94.1  & 94.5 \\ 
     & PointMamba \cite{liang2024pointmamba} & 12.3  & -  & \hspace{1.5pt} 90.2$^\star$ & \hspace{1.5pt} 89.8$^\star$ & 85.4  & 92.4  & - \\ 
    \midrule    
    \multirow{6}[0]{*}{SNN} & KPConv-SNN \cite{wu2024pointsnn} & -  & 40  & -  & -  & 70.5  & -  & 43.9 \\ 
     & Spiking Pointnet \cite{ren2024spiking} & 3.5  & 4   & \hspace{1.5pt} 72.2$^\star$  & \hspace{1.5pt} 76.4$^\star$  & 64.1  & 88.2  & 88.8 \\ 
     & SpikingPointNet \cite{lan2023efficient} & 3.5  & 16  & -  & -  & 69.2  & 88.6  & - \\ 
    % \midrule
     & P2SResLNet-B \cite{wu2024pointsnn} & 14.3 & 1   & \hspace{1.5pt} 78.6$^\star$  & \hspace{1.5pt} 80.2$^\star$  & 74.5  & - & 88.7 \\ 
     & Spiking Point Transformer \cite{wu2025spikingpointtransformerpoint} & 10.2   & 4   & \hspace{1.5pt} 82.8$^\star$  & \hspace{1.5pt} 83.4$^\star$  & 78.0  & 91.4  & \hspace{1.5pt} 91.6$^\star$ \\ 
     & \textbf{Spiking Point Mamba} (\textbf{Ours})  & 12.8  & 4  & \textbf{90.2}(\textcolor{blue}{\small $\uparrow 7.4$})     & \textbf{89.5}(\textcolor{blue}{\small $\uparrow 6.1$})  & \textbf{84.2}(\textcolor{blue}{\small $\uparrow 6.2$})  & \textbf{92.3}(\textcolor{blue}{\small $\uparrow 0.9$}) & \textbf{92.7}(\textcolor{blue}{\small $\uparrow 1.1$}) \\ 
      % & \textbf{Spiking Point Mamba}+\textbf{Pre-training} (\textbf{Ours})  & 12.8  & 4  & \textbf{90.2}(\textcolor{blue}{\small $\uparrow 7.4$})     & \textbf{89.5}(\textcolor{blue}{\small $\uparrow 6.1$})  & \textbf{86.5}(\textcolor{blue}{\small $\uparrow 8.5$})  & \textbf{93.1}(\textcolor{blue}{\small $\uparrow 1.7$}) & \textbf{93.4}(\textcolor{blue}{\small $\uparrow 1.9$}) \\ 
     % & \textit{Improvement} & - & - & \textcolor{blue}{+7.4}   &  \textcolor{blue}{+6.1}  & \textcolor{blue}{+6.2}  & \textcolor{blue}{+0.9}   & \textcolor{blue}{+1.1} \\
    \bottomrule
    \end{tabular}
  % }
    \caption{Performance comparison with the baseline methods. We report overall accuracy (\%) on three ScanObjectNN variants and on ModelNet40 with and without voting. The best results in the SNN domain are bolded, with $\star$ and \textcolor{blue}{\small $\uparrow$} indicating self-reproduced results and the improvement compared to Spiking Point Transformer.}
  \label{tab:main}
\end{table*}

\subsubsection{Datasets}
We evaluate the performance of 3D point cloud classification on the synthetic dataset
ModelNet40~\cite{wu20153d} and the real dataset ScanObjectNN~\cite{uy2019revisiting}.
ModelNet40 contains 40 different object categories, each of
 which contains a large number of 3D object instances.  
ScanObjectNN contains 2902 point clouds from 15 categories. It is a more challenging dataset sampled from real world scans. We follow previous works to conduct experiments on three main variants: OBJ-BG, OBJ-ONLY and PB-T50-RS.

We also evaluate our proposed SPM on the part segmentation dataset ShapeNetPart~\cite{yi2016scalable}, which includes 16,881 models across 16 categories. For pre-training experiments, we utilize ShapeNet~\cite{chang2015shapenet}, a comprehensive dataset containing over 50,000 unique 3D models from 55 common object categories, ensuring robust feature learning.

\subsubsection{Implementation Details}

For classification experiments, the number of input points is set to 2048 by default, adjusted to 1024 for ModelNet40. 
After HDE, the number of points per time step $E$, is set to 256, or 128 for ModelNet40. 
The number of points in the early or late stage is $10\%E$. 
The layer number of Spiking Mamba Block $N$ is 12. 
For most neurons, $V_{th}$ is set to 0.5 while for neurons before each SSM, $V_{th}$  is set to 0.25. 
For segmentation experiments, we utilize the same backbone as in classification experiments. 
For pre-training experiments, the mask ratio is set to 0.6. For fine-tuning experiments, the training epochs is halved to improve training efficiency. 

In most experiments, we utilize the AdamW optimizer and conduct 300 epochs of iterative training on $2\times$RTX 4090 GPUs. The remaining hyperparameters are consistent with previous research~\cite{liang2024pointmamba,han2024mamba3d} for fair comparison.
More experimental details can be found in \cref{sec:Imp}.

\subsection{Experimental Results}
\begin{table}[t]
  \small
  \centering
  \tabcolsep=0.17cm
  % \resizebox{\linewidth}{!}{
    \begin{tabular}{cccc}
    \toprule
    Architecture & Methods & PB-T50-RS & ModelNet40 \\  
    \midrule
    % Example row (replace with actual data)
    \multirow{5}[0]{*}{ANN} & Point-BERT \cite{yu2022point} & 83.1 & 92.7 \\
     & MaskPoint \cite{liu2022masked} & 84.3  & 93.8 \\
     & Point-MAE \cite{pang2022masked} & 85.2  & 93.2 \\
     & Point-M2AE \cite{zhang2022point} & 86.4  & 93.4 \\
     & PointMamba \cite{liang2024pointmamba} & 88.2  & 93.6 \\
    \midrule
    \multirow{2}[0]{*}{SNN}  & SPT\cite{wu2025spikingpointtransformerpoint}$^\star$ & 82.6  & 92.5 \\
     & \textbf{SPM} (\textbf{Ours}) & \textbf{86.5}(\textcolor{blue}{\small $\uparrow 3.9$})  & \textbf{93.1}(\textcolor{blue}{\small $\uparrow 1.4$}) \\
    \bottomrule
    \end{tabular}
  % }
  \caption{Performance of self-supervised learning methods. We report overall accuracy (\%) on PB-T50-RS and ModelNet40.}
  \label{tab:Mae}
\end{table}
\subsubsection{Object Classification}
We compare SPM with other state-of-the-art methods on OBJ-BG, OBJ-ONLY, PB-T50-RS and ModelNet40 datasets and report the results in ~\cref{tab:main} and ~\cref{tab:Mae}.
% It is worth noting that some researches~\cite{qiu2024efficient} based on ILIF~\cite{luo2024integer} have not been discussed in detail, which can be found in Appendix. B. 
All the methods we compare are based on LIF, which is the most popular neuron. Moreover, we use Overall Accuracy (OA) to evaluate the classification performance of all methods.

% On the ScanObjectNN dataset, 从表~\ref{tab:main}中可以看出，我们提出的SPM相对于SNN的基准模型取得了大幅性能提升。具体来说，在SNN领域内，在最难的子数据集PB-T50-RS上，SPM的精度达到了84.2\%，比SPT高6.2\%，在OBJ-ONLY子数据集上，SPM的精度达到了89.5\%，比SPT高-\%，在OBJ-BG子数据集上，SPM的精度达到了90.2\%，比SPT高-\%,这表明了Mamba架构的引入确实能够帮助SPM更好的处理序列建模。同时，在ANN领域内，SPM除了在PB-T50-RS上低于PointMamba 1.2\%，其余子数据集几乎和PointMamba相当，说明SPM在利用Mamba的基础上也利用了SNN的时间特征提取优势，提升了性能。

% On the ModelNet40 dataset,  从表~\ref{tab:main}中可以看出，SPM也取得了SNN领域的SOTA。具体来说，在SNN领域内，SPM的精度达到了92.3\%，比SPT高0.9\%，在ANN领域内，SPM和PointMamba性能相当，但低于PointMLP，考虑到SPM的低能耗优势，SPM在精度和能耗上取得了合理的均衡。

% On the ScanObjectNN dataset, as shown in Table~\ref{tab:main}, our proposed SPM achieves a remarkable performance improvement over the SNN baselines. Specifically, in the SNN domain, 
% On the PB-T50-RS, OBJ-ONLY, and OBJ-BG subsets, SPM attains 84.2\%, 89.5\%, and 90.2\% OA, reflecting a 6.2\%, 6.1\%, 7.4\% improvement over SPT.
% This means that the introduction of Mamba significantly boosts SPM's sequence modeling capabilities.
% In the ANN domain, SPM performs similarly to PointMamba, with a slight reduction in OA on PB-T50-RS. This means that in addition to leverage Mamba, SPM also benefits from the temporal feature extraction of SNN.

% On the ModelNet40 dataset, as shown in Table~\ref{tab:main}, SPM also achieves the state-of-the-art performance in the SNN domain. 
% Specifically, in the SNN domain, SPM attains 92.3\% OA, reflecting a 0.9\% improvement over SPT. In the ANN domain, SPM performs similarly to PointMamba. 
As shown in ~\cref{tab:main}, our proposed SPM achieves the state-of-the-art performance in the SNN domain on both ScanObjectNN and ModelNet40 datasets, significantly outperforming SNN baselines. 
Specifically, on the PB-T50-RS, OBJ-ONLY, and OBJ-BG subsets, SPM attains 84.2\%, 89.5\%, and 90.2\% OA, surpassing SPT by 6.2\%, 6.1\%, and 7.4\%, respectively. Moreover, on the ModelNet40 dataset, SPM achieves 92.3\% OA, reflecting a 0.9\% improvement over SPT. 
This means that the introduction of Mamba enhances SPM’s sequence modeling capabilities.
In the ANN domain, SPM performs comparably to PointMamba, confirming that SPM benefits not only from Mamba but also from the temporal feature extraction of SNN.

\begin{table}[t]
\small
  \centering
  \tabcolsep=0.17cm
  % \resizebox{\linewidth}{!}{
    \begin{tabular}{cccc}
    \toprule
    Architecture & Methods & Cls. mIoU & Ins. mIoU \\
    \midrule
    \multirow{5}[0]{*}{ANN} & PointNet~\cite{qi2017pointnet}& 80.39 & 83.7 \\
     & PointNet++~\cite{qi2017pointnet++} & 81.85 & 85.1 \\
     & DGCNN~\cite{phan2018dgcnn}  & 82.33 & 85.2 \\
     & APES~\cite{wu2023attention} & 83.67 & 85.8\\
     & PointMamba \cite{liang2024pointmamba} & 84.29 & 85.8 \\
    \midrule
    \multirow{2}[0]{*}{SNN}  & SPT\cite{wu2025spikingpointtransformerpoint}$^\star$ & 81.32  & 82.9 \\
     & \textbf{SPM} (\textbf{Ours}) & \textbf{82.29}(\textcolor{blue}{\small $\uparrow 0.97$})  & \textbf{84.8}(\textcolor{blue}{\small $\uparrow 1.9$}) \\
     % & \textbf{SPM}+\textbf{Pre-training} & \textbf{82.29}(\textcolor{blue}{\small $\uparrow 0.97$})  & \textbf{84.8}(\textcolor{blue}{\small $\uparrow 1.9$}) \\
    \bottomrule
    \end{tabular}
  % }
  \caption{Performance of part segmentation methods. We report Cls. mIoU (\%) and Ins. mIoU (\%) on ShapeNetPart.}
  \label{tab:Partseg}
\end{table}
\begin{figure*}[t]
    \centering
    \includegraphics[width=1.0\linewidth]{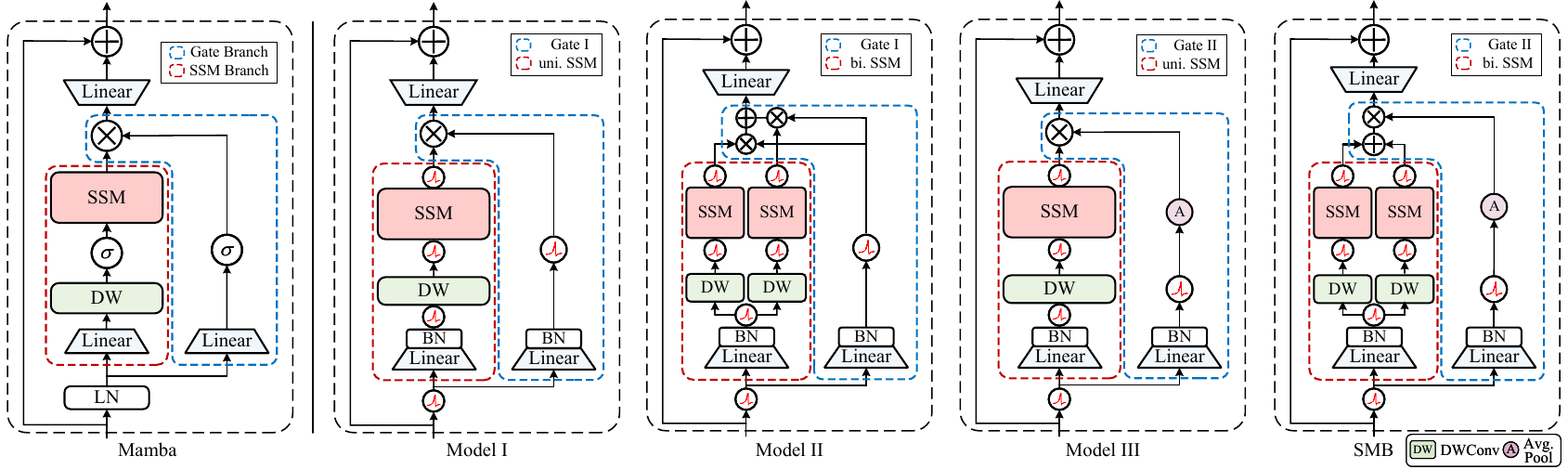}
    \caption{Ablation study of the design of SMB. Model I-III denote three variant models of SMB while Mamba denotes ANN counterpart. Gate I and Gate II represent vanilla and SNN-adapted Gate Branch. uni. and bi. SSM represent unidirectional and directional SSM.}
    \label{fig:SMB}
\end{figure*}

% \subsubsection{Pre-traing and Finetune} 

% In the self-supervised learning task, 
As shown in ~\cref{tab:Mae}, we also evaluate the performance of spike-based pre-training model for SPM using two representative datasets: PB-T50-RS and ModelNet40. In the SNN domain, our SPM achieves the state-of-the-art performance and outperforms SPT with 86.5\% OA on PB-T50-RS and 93.1\% OA on ModelNet40. 
In addition, compared to training from scratch on PB-T50-RS and ModelNet40, SPM shows an improvement of 2.3\%, and 0.8\% respectively.
In the ANN domain,  SPM performs nearly on par with Point-M2AE. These results illustrate that spike-based pre-training effectively enhances the SNN encoder.

\begin{table*}[t!]
  \small
  \centering
  \tabcolsep=0.29cm
  % \resizebox{\linewidth}{!}{
    \begin{tabular}{cccccccccc}
    \toprule
    \multicolumn{1}{c}{\multirow{2}[2]{*}{Architecture}} & 
    \multicolumn{1}{c}{\multirow{2}[2]{*}{Models}} & 
    \multicolumn{2}{c}{Gate Branch} & 
    \multicolumn{2}{c}{SSM Branch} & 
    \multicolumn{3}{c}{ScanObjectNN} & 
    \multicolumn{1}{c}{\multirow{2}[2]{*}{ModelNet40}} 
    \\ 
    \cmidrule(lr){3-4}
    \cmidrule(lr){5-6}
    \cmidrule(lr){7-9}
     & & Gate I & Gate II & bi. & uni. &  OBJ-BG & OBJ-ONLY & PB-T50-RS   \\
     \midrule
     ANN & Mamba & - & - & - & - & 90.2 & 89.5 & 84.2 & 92.3   \\
     \midrule
     \multirow{4}[2]{*}{SNN} & Model I & \ding{51} & \ding{55} & \ding{55} & \ding{51} & 88.3 & 87.9 & 83.1 & 91.4\\
     & Model II  & \ding{51} & \ding{55} & \ding{51} & \ding{55} & 89.3 & 88.9 & 83.8 & 91.8\\
     & Model III  & \ding{55} & \ding{51} & \ding{55} & \ding{51} & 88.9 & 88.4 & 83.5 & 91.6  \\
     & SMB & \ding{55} & \ding{51} & \ding{51} & \ding{55} & \textbf{90.2}     & \textbf{89.5}  & \textbf{84.2}  & \textbf{92.3} \\

    \bottomrule
    \end{tabular}
  % }
  \caption{Ablation study of the design of SMB. We report overall accuracy (\%) on three ScanObjectNN variants and ModelNet40.}
  \label{tab:SMB}
\end{table*}

\subsubsection{Part Segmentation} 

% We further evaluate SPM on the ShapeNetPart dataset using Cls. mIoU and Ins. mIoU, which represent mean intersection over union for categories and instances respectively.  As shown in ~\cref{tab:Partseg}, in the SNN domain, our proposed SPM achieves  superior performance, with 82.29\% Cls. mIoU and 84.8\% Ins. mIoU, surpassing SPT by 1.9\% in Ins. mIoU and 0.97\% in Cls. mIoU. In the ANN domain,  while SPM slightly lags behind PointMamba,
% it provides a compelling trade-off between performance and energy efficiency. From \cref{tab:time}, it can be observed that the energy consumption of SPM 远远低于 PointMamba。

We further evaluate SPM on the ShapeNetPart dataset using two metrics: Cls. mIoU and Ins. mIoU, which measure the mean intersection over union for categories and instances, respectively. As shown in ~\cref{tab:Partseg}, SPM demonstrates superior performance within the SNN domain, achieving 82.29\% Cls. mIoU and 84.8\% Ins. mIoU, surpassing SPT by 0.97\% and 1.9\%, respectively. In the ANN domain, while SPM slightly lags behind PointMamba in mIOU, it offers a compelling balance between performance and energy efficiency as further illustrated in ~\cref{tab:time}.

% We further evaluate SPM on the ShapeNetPart dataset using two key metrics: Cls. mIoU and Ins. mIoU, which measure the mean intersection over union for categories and instances, respectively. As shown in ~\cref{tab:Partseg}, SPM demonstrates superior performance within the SNN domain, achieving 82.29\% Cls. mIoU and 84.8\% Ins. mIoU, surpassing SPT by 1.9\% in Ins. mIoU. In the ANN domain, while SPM slightly trails behind PointMamba, it offers a compelling trade-off between performance and energy efficiency. By leveraging spike-based biological characteristics, SPM achieves a balance between computational cost and accuracy, making it a promising alternative for energy-efficient 3D processing.

\subsection{Ablation Studies}

To analyze each module in SPM, we conduct a series of ablative experiments on ScanObjectNN and ModelNet40 datasets. We construct three variants Model I-III, which correspond to the ablation of our proposed improvements in two branches. Furthermore, we conduct ablation studies on the bidirectional strategy, input encoding and time step.

% The former introduces temporal dynamics while preserving key information at each time step. The latter enhances sample diversity in the later stage, reducing redundancy and even noise interference.

\subsubsection{Ablation on SMB} 

Our proposed SMB enhances both Gate and SSM branches of Mamba. Therefore, we conduct ablation studies on them, constructing three variant models (Model I-III) as shown in \cref{fig:SMB}. Additionally, we conduct a more detailed ablation on the bidirectional strategy of the SSM branch.

\noindent \textbf{Branch Architecture}
\cref{fig:SMB} shows how SPM enhances Mamba’s gating and SSM branches for better adaptation to SNNs. Specifically, Gate I merely replaces ReLU/SiLU with LIF, while Gate II minimizes information loss from spike gating. The uni. and bi. refer to unidirectional and bidirectional SSM branches.
From \cref{tab:SMB}, we observe that SPM outperforms Models I-III. Compared to Model I, SPM achieves improvements of 1.9\%, 1.6\%, 1.1\%, and 0.9\% OA on OBJ-BG, OBJ-ONLY, PB-T50-RS, and ModelNet40, respectively. This indicates that simply applying Mamba to SNNs is insufficient. Additionally, both Model II and Model III outperform Model I, highlighting the effectiveness of the two SNN-adaptive branches. With these branches, SMB performs similarly to Mamba while reducing energy consumption.
\begin{table}[t!]
    \centering
    \small
    \resizebox{\columnwidth}{!}{
        \begin{tabular}{ccccc}
        \toprule
        \multicolumn{2}{c}{SSM Branch} & \multirow{2}[2]{*}{OBJ-BG} & \multirow{2}[2]{*}{PB-T50-RS} & \multirow{2}[2]{*}{ModelNet40} \\ 
        \cmidrule(lr){1-2}    
        Token Flip & Time Flip & & & \\
        \midrule
        \ding{55} & \ding{55} & 88.9 & 83.5 & 91.6 \\
        \ding{51} & \ding{55} & 89.6 & 83.7 & 91.7 \\
        \ding{51} & \ding{51} & 89.2 & 83.4 & 91.6 \\
        \ding{55} & \ding{51} & \textbf{90.2} & \textbf{84.2} & \textbf{92.3} \\
        \bottomrule
        \end{tabular}
    }
    \caption{Ablation study of the bidirectional strategy. 
    We report overall accuracy (\%) on OBJ-BG, PB-T50-RS and ModelNet40. }
    \label{tab:flip}
\end{table}

% 不同于传统双向Mamba结构,SMB因为点云无序性的固有特性,取消了Token Flip也就是反转Token维度, 来防止网络习得不稳定的伪序列依赖,而使用 Time Flip 仅仅反转时间维度来促使SSM学习时间交互.
% 在表~\ref{tab:flip}中，我们对这两种双向策略进行了消融实验。结果显示，当这两种策略都不使用时，SSM分支退化为单向SSM，性能较差，这与SSM的单向建模特性有关。当仅使用Token Flip时，相比于单向SSM，精度有所提升。然而，与仅使用Time Flip时相比，Token Flip在所有数据集上分别表现出0.4%、0.4%和0.3%的精度下降。值得注意的是，当两种策略同时使用时，模型性能反而退化至与单向SSM相当。这表明，在SMB的设计中，双向策略至关重要。具体而言，Time Flip比Token Flip更能促使网络学习到深层次的交互信息，但过于复杂的双向策略可能会导致性能下降，这可能是由于输入SSM的脉冲矩阵信息的限制所致。
\noindent \textbf{Bidirectional Strategy}
In ~\cref{tab:flip}, we further evaluate the impact of two bidirectional strategies.
Token Flip reverses the token dimension, while Time Flip reverses the time dimension.
When neither is used, SSM branch degenerates into unidirectional SSM leading to poor performance.
Using only Token Flip performs better than unidirectional SSM, but reflects a 0.4\%, 0.4\%, and 0.3\% drop in OA compared to Time Flip. Additionally, applying both strategies degrades performance to that of unidirectional SSM.
These results highlight the critical role of bidirectional strategies in SMB. Specifically, Time Flip proves more effective than Token Flip in enabling the network to capture more meaningful interaction. 
However, the combination of both strategies hinders model performance, likely due to the limited information in the spike input to SSM.

% In Table~\ref{tab:flip}, we evaluate the impact of two bidirectional strategies. Token Flip reverses the token dimension, while Time Flip reverses the time dimension. Without either, the SSM branch degenerates into a unidirectional SSM, leading to poor performance.

% Using only Token Flip improves performance over the unidirectional SSM but results in a 0.4%, 0.4%, and 0.3% OA drop across all datasets compared to Time Flip. Applying both strategies together degrades performance to the unidirectional level, likely due to the limited spike input to SSM. These results underscore the importance of bidirectional strategies, with Time Flip proving more effective in capturing meaningful interactions.

\begin{table}[t!]
    \small
    \centering
    \resizebox{\columnwidth}{!}{
        \begin{tabular}{ccccc}
        \toprule
        \multicolumn{2}{c}{HDE} & \multirow{2}[2]{*}{OBJ-BG} & \multirow{2}[2]{*}{PB-T50-RS} & \multirow{2}[2]{*}{ModelNet40} \\ 
        \cmidrule(lr){1-2}       
        Forward & Backward & & &  \\ 
        \midrule
        \ding{55} & \ding{55} & 88.6& 83.7& 91.5  \\
        \ding{55} & \ding{51} & 89.8& 83.8& 92.0  \\
        \ding{51} & \ding{55} & 89.5& 84.0& 91.7  \\
        \ding{51} & \ding{51} & \textbf{90.2} & \textbf{84.2} & \textbf{92.3}  \\
        \bottomrule
        \end{tabular}
    }
    \caption{Ablation study of encoding method performance. We
    report overall accuracy (\%) on OBJ-BG, PB-T50-RS and ModelNet40. Forward denotes \textit{finite forward sliding} while Backward denotes \textit{infinite backward extension}.}
    \label{tab:HDE}
\end{table}

\subsubsection{Ablation on HDE}
% 缺乏对 x x 直接编码 的解释

As shown in \cref{tab:HDE}, we evaluate the impact of \textit{finite forward sliding} and \textit{infinite backward extension} of HDE 
 on the OBJ-BG, PB-T50-RS and ModelNet40 datasets.
When both are applied simultaneously, SPM achieves the best performance on all datasets, with 90.2\%, 84.2\%, and 92.3\% OA respectively. 
Conversely, when neither of them is used, HDE degenerates into direct encoding, reflecting a relative decline in performance. Furthermore, the performance also improves when either of them is used individually.
This means that  both \textit{finite forward sliding} and \textit{infinite backward extension} contribute effectively to model performance.

\begin{table}[t!]
    \centering
    \resizebox{\columnwidth}{!}{
        \begin{tabular}{ccccc}
        \toprule
        Timestep & Energy(mJ) & OBJ-BG & PB-T50-RS & ModelNet40   \\
        \midrule
        ANN & 18.9 & 90.2 & 85.4 & 92.4 \\
        \midrule
        1 & \textbf{1.5} & 88.9 & 83.3 & 91.6 \\
        2 & 2.8 & 89.8 & 83.7 & 91.8 \\
        3 & 3.9 & 90.2 & 83.8 & 92.1 \\ 
        4 & 5.4 & \textbf{90.2} & 84.2 & \textbf{92.3} \\
        6 & 7.6 & 90.0 & \textbf{84.3} & 92.3 \\
        \bottomrule
        \end{tabular}
    }
    \caption{Ablation study of time step. We
    report overall accuracy (\%) and Energy(mJ) on OBJ-BG, PB-T50-RS and ModelNet40. ANN refers to PointMamba.}
    \label{tab:time}
\end{table}

\subsubsection{Ablation on Time Step} 

In \cref{tab:time}, we further evaluate SPM’s performance and energy consumption on OBJ-BG, PB-T50-RS and ModelNet40 under different time steps, including PointMamba as ANN counterpart for comparison. Given the high computational cost, the maximum time step is set to 6.  

Unlike previous works~\cite{wu2024pointsnn,ren2024spiking}, we can see that in \cref{tab:time}, SPM’s performance improves with time steps, plateauing at 4 and 6 with 90.2\%, 84.3\%, and 92.3\% OA, respectively. This suggests that by introducing temporal dynamics and leveraging SNNs’ time feature extraction, SPM alleviates the redundancy of direct encoding in point cloud analysis~\cite{wu2024pointsnn}.
Meanwhile, SPM achieves a 12.6$\times$ energy reduction over PointMamba at 1 time step and 3.5$\times$ at 4 time steps while maintaining comparable performance. 
% Considering all factors, we report results at 4 time steps in ~\cref{tab:main}.

\begin{figure}[t]
    \centering
    \includegraphics[width=1.0\linewidth]{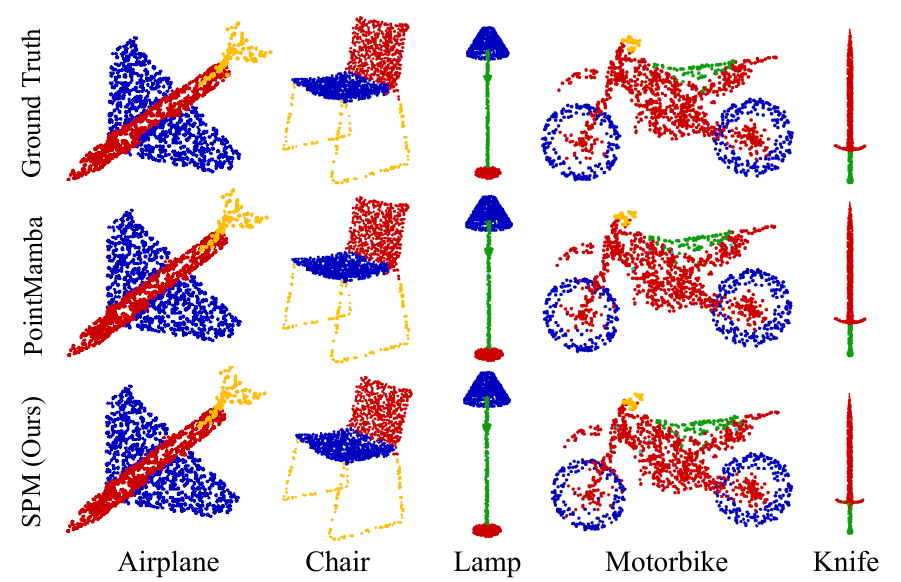}
    \caption{Qualitative results of part segmentation of our SPM and ANN counterpart (PointMamba) on ShapeNetPart.}
    \label{fig:Partseg}
\end{figure}

\begin{figure}[t]
    \centering
    \includegraphics[width=1.\linewidth]{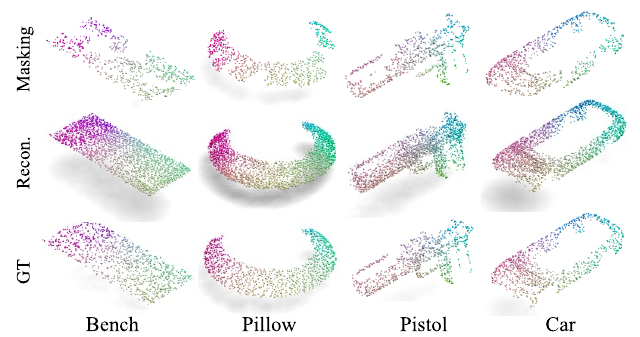}
    \caption{Qualitative results of reconstruction on ShapeNet.}
    \label{fig:reconstruction}
\end{figure}

\subsection{Visualizations}

We visualize part segmentation results of SPM and its ANN counterpart, PointMamba, in ~\cref{fig:Partseg}. The segmentation visualizations of SPM are nearly identical to PointMamba, with minor differences in fine details, while overall quality remains intact. Additionally, we present reconstruction visualizations of the pre-training in ~\cref{fig:reconstruction}. Despite the high masking ratio, SPM effectively reconstructs the overall object shape. More visualizations can be seen in \cref{sec:Visualizations}.

% We visualize part segmentation results of SPM and its ANN counterpart, PointMamba, in ~\cref{fig:Partseg}. The segmentations visualizations of them are nearly identical, with minor differences in fine details, while overall quality remains intact. Reconstruction visualizations of the pre-train mask modeling are shown in ~\cref{fig:reconstruction}. Despite a high masking ratio, SPM effectively reconstructs the object's shape. More visualizations are in \cref{sec:Visualizations}.

\section{Conclusion}

In this work, we present Spiking Point Mamba (SPM), a novel SNN framework based on Mamba architecture for point cloud analysis. 
The proposed HDE and SMB effectively introduce temporal dynamics and utilize the modeling capacity of Mamba and SNNs, achieving a remarkable performance improvement in the SNN domain. 
We hope that SPM can further reduce energy consumption in future research for neuromorphic hardware deployment.

{
    \small
    \bibliographystyle{ieeenat_fullname}
    \bibliography{main}
}

\clearpage
\setcounter{page}{1}
\maketitlesupplementary
\appendix

In the supplementary material, we further explore the behavior of Farthest Point Sampling (FPS) and experimentally validate the theoretical foundation of HDE. We present more ablation studies on neurons and mask ratios, as well as comparing our SPM with additional models~\cite{qiu2024efficient}. Next, we provide more implementation details of our SPM. Finally, we include additional visualizations across various tasks.

\section{Farthest Point Sampling Behavior}
\label{sec:behavior}

Farthest Point Sampling (FPS) is a widely used technique  for point cloud analysis, which select a subset of points from a larger point cloud in a way that maximizes the minimum distance between selected points. FPS can be divided into three key stages: early, middle, and late stages.

\noindent \textbf{Early Stage.}
The early stage of FPS involves random initial selection, which can cause instability in the sampling process. The points selected in this stage are heavily influenced by the random choice, leading to a not so well distribution across the point cloud. This randomness results in high variance, making the early stage less reliable for capturing the overall structure of the point cloud. Together, the early stage is unstable due
to random initial selection.

\noindent \textbf{Middle Stage.} 
As the algorithm progresses, the middle stage becomes more stable. FPS starts to cover important features of the geometry, improving the distribution of sampled points. The algorithm focuses on regions with significant features while maintaining a good spread across the point cloud. This stability enables a more meaningful representation of the point cloud. Together, the middle stage stabilizes and captures the skeletal structure.

\noindent \textbf{Late Stage.} In the late stage, FPS starts to experience diminishing returns. As the number of selected points increases, the algorithm starts to introduce redundancy or noise. The points selected in this stage are often located in areas that are already well-represented by previous selections, leading to overlapping regions.Together, the late stage may introduce redundancy or noise.
\begin{table*}[t]
\setlength{\tabcolsep}{3.4mm}
\centering
\small
\tabcolsep=0.55cm
\begin{tabular}{lcccc}
 \toprule
\multirow{2.3}{*}{Configuration}  & Pre-training  & \multicolumn{2}{c}{Classification}   & Segmentation \\
\cmidrule(r){3-4}
& ShapeNet & ModelNet40 & ScanObjectNN & ShapeNetPart \\
 \midrule
 Optimizer & AdamW & AdamW  & AdamW & AdamW\\
 Learning rate & 1e-3 & 1e-3 & 1e-3 & 2e-4 \\
 Weight decay & 5e-2 & 5e-2 & 5e-2  & 5e-2 \\
 Learning rate scheduler & cosine &cosine & cosine & cosine \\
 Training epochs & 300 & 300 & 300 & 300\\
 Warmup epochs & 20 & 30 & 30 & 20\\
 Batch size & 128 & 72 & 36 & 36\\
 \midrule
Num. of encoder layers $N$ & 12 &12 &12 & 12\\
Num. of decoder layers $N_d$ & 4 & -& -& -\\
Input points $M$ & 1024 & 1024 & 2048 & 2048\\
Num. of patches $n$& 128 & 128& 256 &256\\
Patch size $k$ & 32 &32 & 32 &32\\
 \midrule
 Augmentation & Scale\&Trans & Scale\&Trans & Rotation & - \\
\bottomrule
\end{tabular}
\caption{Implementation details for pre-training and downstream tasks such as classfication and segmentation tasks.}
\label{tab:hyper_params}
\end{table*}

\begin{figure}[t]
    \centering
    \includegraphics[width=1.0\linewidth]{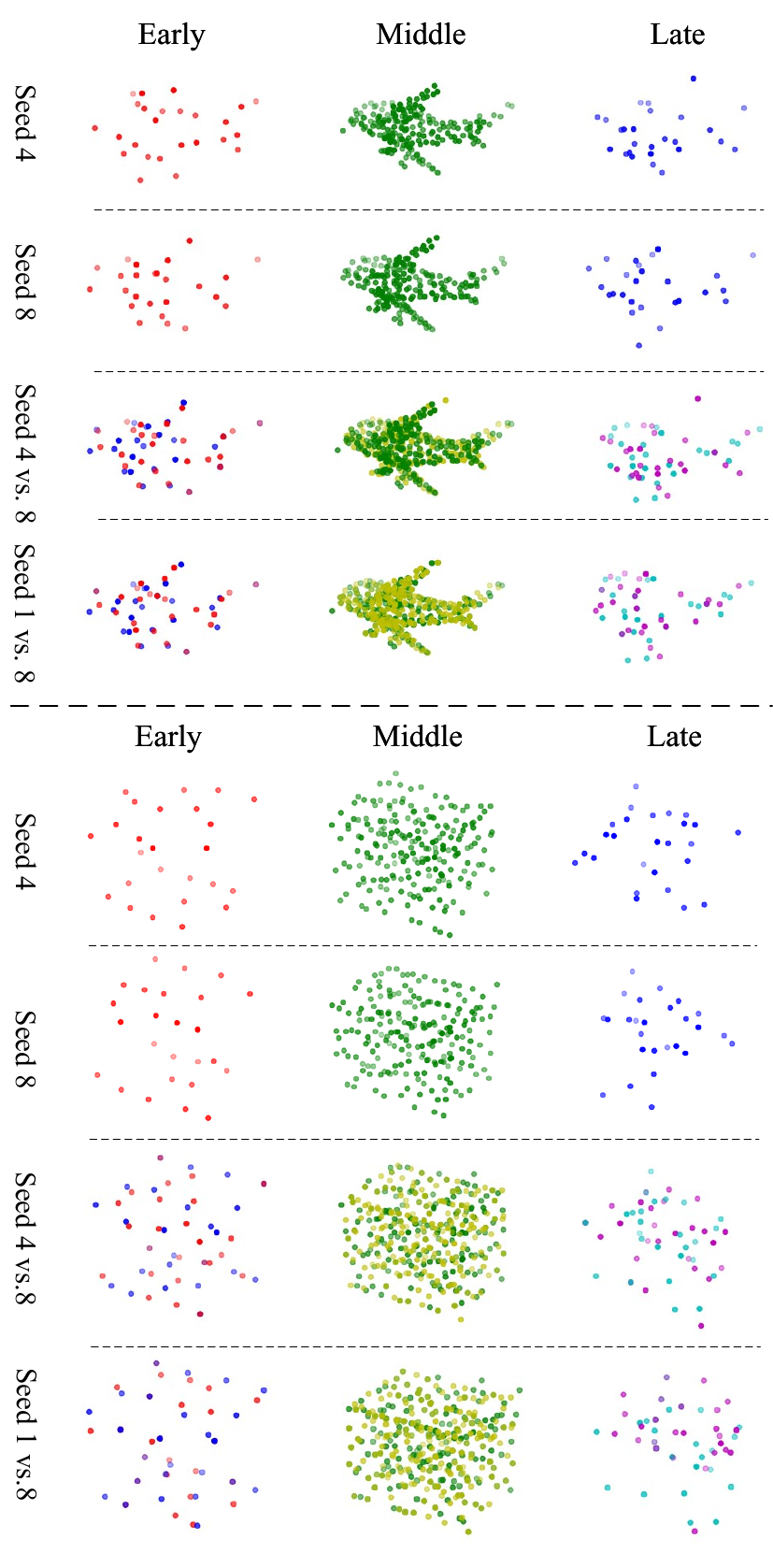}
    \caption{Quantitative comparison of the sampling structures of the early, middle, and late stages under different random seeds.}
    \label{fig:FPS}
\end{figure}

We use Chamfer Distance to measure the similarity between the early, middle, and late stages under different random seeds in ~\cref{tab:fps}, and visualize the sampling structures of the early, middle, and late stages under certain random seeds for quantitative comparison in ~\cref{fig:FPS}, thereby demonstrating the rationale behind the hierarchical stages.

From ~\cref{fig:FPS}, we can see that in the early stage under different random seeds, FPS is unstable and does not fully represent the object information, indicating a strong influence from the random initial points. In the middle stage, the points show little visual change and provide a stable representation of the entire point cloud. However, in the late stage, the point cloud no longer captures the overall structure, exhibiting a highly variable shape, and can only be considered as redundant points.

From ~\cref{tab:fps}, we can observe that the Chamfer Distance between the early, middle, and late stage point clouds under different random seeds can also be understood as their similarity. Multiple experiments indicate that the early and late stage points exhibit more unstable distributions compared to the middle stage points, which can be intuitively seen from the visualization in ~\cref{fig:FPS}. Furthermore, the middle stage points show a high similarity across different random seeds, confirming that they reliably represent the skeletal structure of the entire point cloud. Additionally, we observe that the similarity between the late and middle stage points is also high, suggesting an overlap between these two stages and further supporting the redundancy characteristic of the late stage points.

\section{Ablation Study}
The ablation study mainly supplements performance comparison experiments for SPM with different neurons, as well as comparisons with other models~\cite{qiu2024efficient} using ILIF~\cite{luo2024integer} and its training strategies. Additionally, we conduct an ablation study on the pre-training mask ratio.

\noindent \textbf{Ablation on different neurons.}
In ~\cref{tab:neurons}, we conduct a detailed ablation study using different neurons on the OBJ-BG, OBJ-ONLY, PB-T50-RS and ModelNet40 datasets. It can be observed that the accuracy of traditional IF neurons and their variants tends to fluctuate only slightly, whereas ILIF significantly improves the model’s performance. This improvement is due to its integer-based training and spike-based inference mechanism. Therefore, a direct comparison with the traditional LIF model is not entirely reasonable. Here, we present the experimental results of SPM using ILIF, where accuracy reaches 91.5\%, 91.2\%, and 85.2\% on OBJ-BG, OBJ-ONLY and PB-T50-RS datasets respectively, and 93.0\% on the ModelNet40 dataset.
\begin{table}[t!]
    \centering
    \resizebox{\columnwidth}{!}{
        \begin{tabular}{ccccc}
        \toprule
        \multirow{2}[2]{*}{Seed} &  \multicolumn{4}{c}{Chamfer Distance} \\ 
        \cmidrule(lr){2-5}       
        & $D^{Early,Early}$ &  $D^{Mid,Mid}$ &  $D^{Mid,Late}$ & $D^{Late,Late}$   \\
        \midrule
        1 vs. 2 & 0.18 & 0.09 & 0.11 & 0.24 \\
        4 vs. 7 & 0.17 & 0.07 & 0.09 & 0.20 \\
        6 vs. 25 & 0.16 & 0.06 & 0.08 & 0.22 \\
        16 vs. 32 & 0.19 & 0.06 & 0.10 & 0.25 \\ 
        44 vs. 42 & 0.18 & 0.06 & 0.10 & 0.22 \\ 
        123 vs. 321 & 0.19 & 0.06 & 0.07 & 0.23 \\ 
        
        \midrule
        \textit{Mean} & \textcolor{blue}{0.18} & \textcolor{blue}{0.07} & \textcolor{blue}{0.09} & \textcolor{blue}{0.23} \\
        \bottomrule
        \end{tabular}
    }
    \caption{The similarity of the early, middle, and late stages under different random seeds. ($a$ vs. $b$, $D^{A,B}$) denotes the Chamfer distance between stage $A$ under random seed $a$ and stage $B$ under random seed $b$.}
    \label{tab:fps}
\end{table}

\begin{table}[t!]
    \centering
    \resizebox{\columnwidth}{!}{
        \begin{tabular}{cccccc}
        \toprule
        \multirow{2}[2]{*}{Method} &  \multirow{2}[2]{*}{Neurons} &  \multicolumn{3}{c}{ScanObjectNN} & \multirow{2}[2]{*}{ModelNet40}\\ 
        \cmidrule(lr){3-5}       
        & & OBJ-BG & OBJ-ONLY & PB-T50-RS &    \\
        \midrule
        E-3DSNN & ILIF~\cite{luo2024integer} & 86.5$^\star$ & 86.0$^\star$ & 80.4$^\star$ & 91.7\\
        \midrule
        
        \multirow{5}[2]{*}{SPM} & IF~\cite{bulsara1996cooperative} & 89.8 & 89.0 & 84.1 & 92.0 \\
        & LIF~\cite{gerstner2002spiking} & 90.2 & 89.5 & 84.2 & 92.3  \\
        & EIF~\cite{brette2005adaptive} & 89.9 & 89.2 & 84.1 & 92.1 \\
        & PLIF~\cite{fang2021incorporating} & 90.5 & 89.6 & 84.1 & 92.5 \\ 
        & ILIF~\cite{luo2024integer} & \textbf{91.5} & \textbf{91.2} & \textbf{85.2}& \textbf{93.0} \\ 
        & \textit{Improvement} &  \textcolor{blue}{+5.0}   &  \textcolor{blue}{+5.2}  & \textcolor{blue}{+4.8}  & \textcolor{blue}{+1.3}\\
        \bottomrule
        \end{tabular} 
    }
    \caption{ Ablation study on different neurons with 4 time steps for IF, LIF, EIF and PLIF and $1\times 4$ for ILIF.}
    \label{tab:neurons}
\end{table}
\noindent \textbf{Comparison with other models.}
In ~\cref{tab:neurons}, we also make an additional comparison between SPM and other ILIF-based models such as E-3DSNN. To ensure a fair comparison, we used a configuration of $T \times D$ as $1 \times 4$, which follows the ILIF training strategy, allowing the model to convert into 4 time steps for spike-based inference during the inference phase.
In the case of using ILIF equally, we observe that our SPM significantly outperforms benchmark models such as E-3DSNN, with overall accuracy reaching 91.5\%, 91.2\%, 85.2\%, and 93.0\% on OBJ-BG, OBJ-ONLY, PB-T50-RS, and ModelNet40, respectively.

\noindent \textbf{Ablation on mask ratios}
In ~\cref{tab:mask-ratio}, we conduct a detailed ablation study on the masking ratio. 
It can be observed that when the masking ratio is set to 0.6, which is the ratio we ultimately selected, the accuracy of the fine-tuning task reaches their optimal performance, outperforming other settings. 
This suggests that for spike-based pre-training, too much masking can lead to excessive information loss, preventing the SNN encoder from learning meaningful feature representations. On the other hand, too little masking may reduce the difficulty of the reconstruction task, also causing SNN encoder to fail to learn strong feature representations.

\section{Implement details}
\label{sec:Imp}
\begin{table}[t!]
\small
\tabcolsep=0.4cm
    \centering
	\begin{tabular}{ccccc}
        \toprule
        Masking ratio& Loss & PB-T50-RS & ModelNet40\\
        \midrule
        0.4 & 1.66 & 86.1& 92.6   \\
        0.6 & \textbf{1.57} & \textbf{86.5} & \textbf{93.1}  \\
        0.8 & 2.03 & 85.9 & 92.7  \\
        0.9 & 2.00 & 86.0 & 92.5  \\
        \bottomrule
        \end{tabular}
    \caption{Ablation study on masking strategy. The pre-training loss (× 1000) along with fine-tuning accuracy (\%) are reported on PB-T50-RS and ModelNet40.}
    \label{tab:mask-ratio}
\end{table}
In this section, we provide more specific details about the training parameters for each dataset, as shown in ~\cref{tab:hyper_params}. Different training hyperparameters were used for different datasets, but the backbone remained the same, consisting of a 12-layer stacked SPM, which facilitated the generalization across different datasets and the fine-tuning of pre-training models.
\section{More Visualizations}
\label{sec:Visualizations}

\begin{figure}[t]
    \centering
    \includegraphics[width=1.0\linewidth]{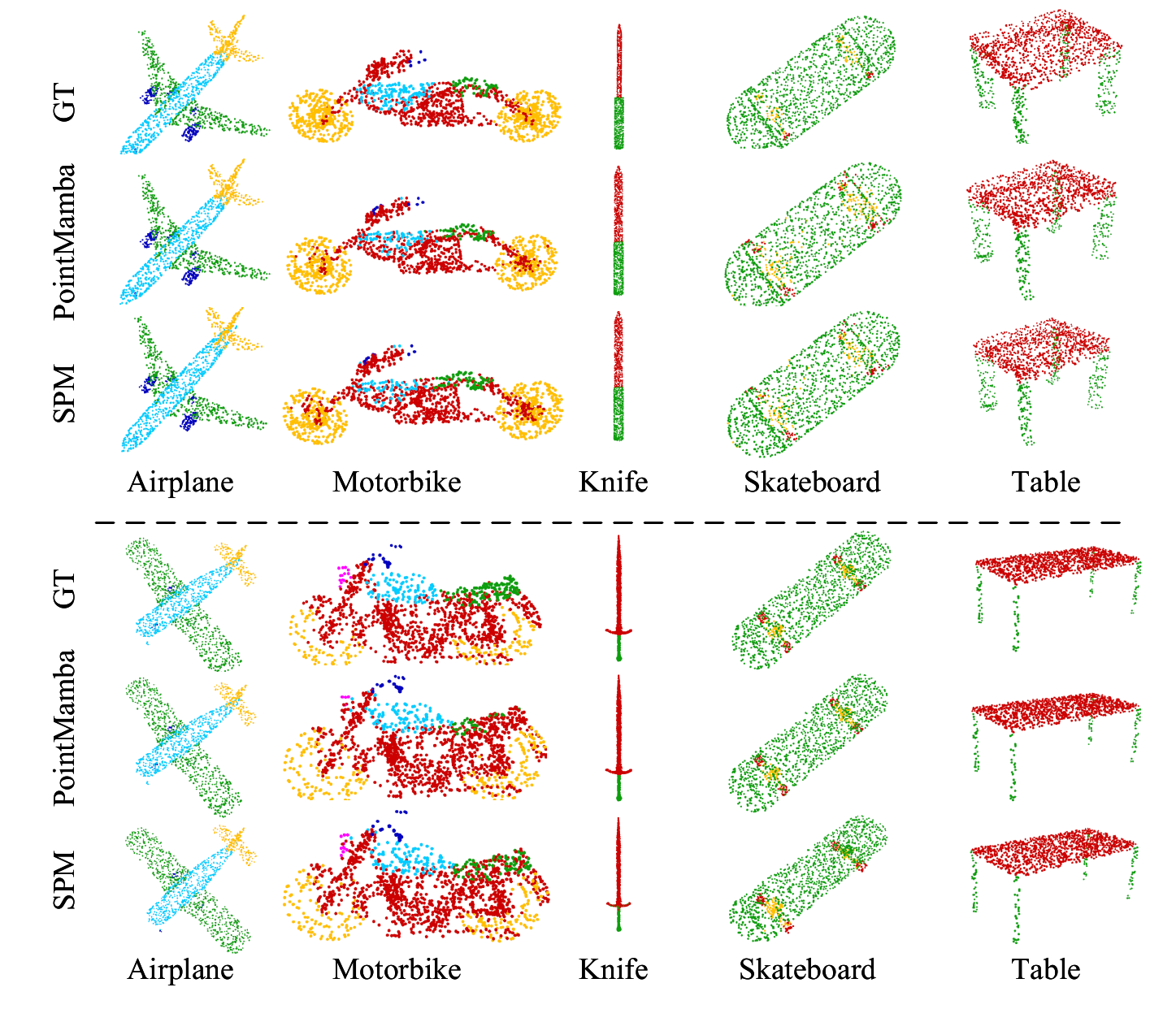}
    \caption{Qualitative results of part segmentation of our SPM and ANN counterpart (PointMamba) on ShapeNetPart.}
    \label{fig:Partseg_supp}
\end{figure}

\begin{figure}[t]
    \centering
    \includegraphics[width=1.0\linewidth]{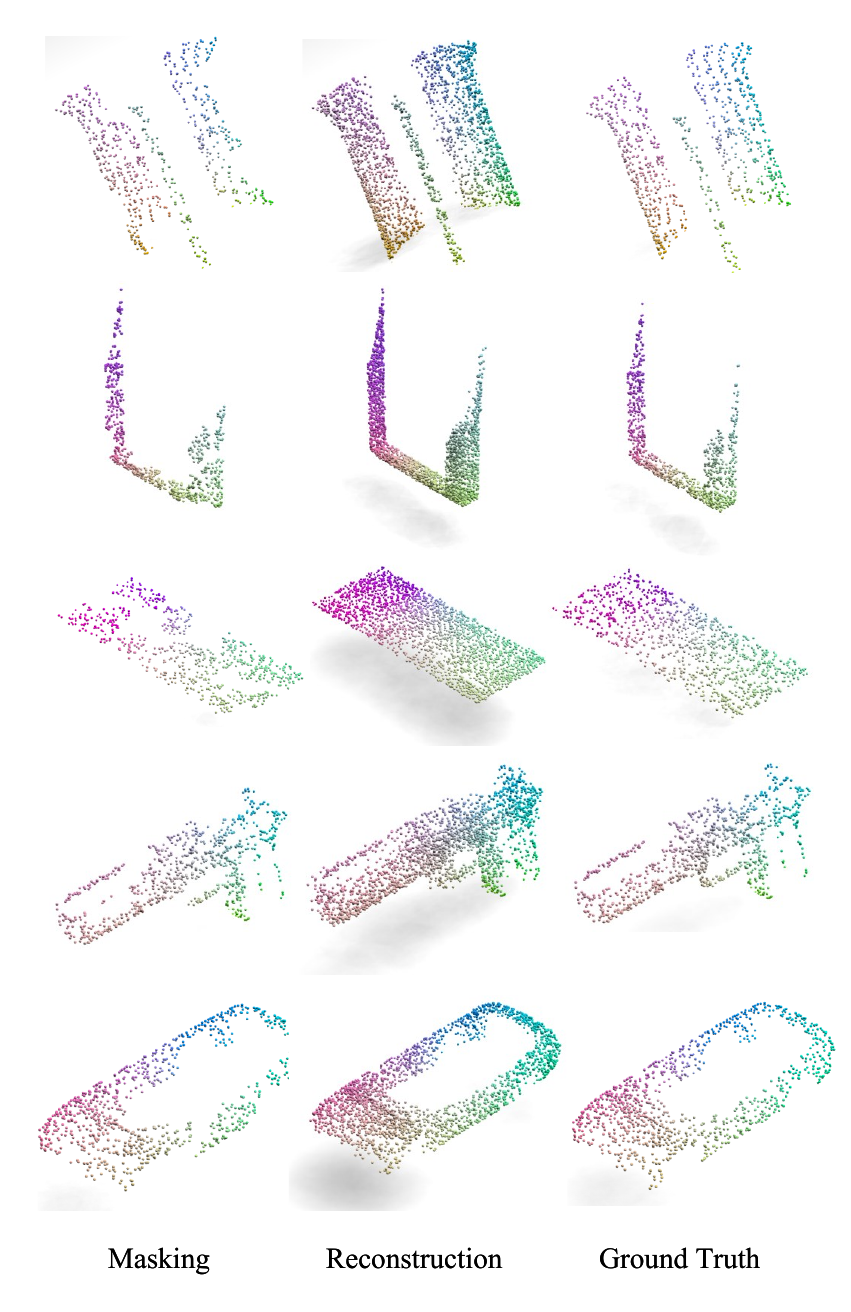}
    \caption{Qualitative results of reconstruction on ShapeNet.}
    \label{fig:reconstruction_supp}
\end{figure}

The figure of main paper shows only a few sample visualizations. In this section, we provide more visualizations to quantitatively demonstrate the effectiveness and performance of our SPM, as shown in \cref{fig:Partseg_supp} and \cref{fig:reconstruction_supp}.

As can be seen from ~\cref{fig:Partseg_supp}, the part segmentation results of SPM are almost identical to those of PointMamba across different classes. Slightly more complex classes may show a minor difference, but it does not significantly affect the overall performance. From ~\cref{fig:reconstruction_supp}, it can also be observed that our SPM performs excellently in the pre-trained reconstruction task. Despite the large masking range, it is still able to recover the general shape of the object.

\end{document}